\documentclass{jdmdh}
\usepackage{array}
\usepackage{pgfplots}
\pgfplotsset{compat=1.18}

\usepackage[english]{babel}
\usepackage[utf8]{inputenc}
\usepackage{natbib}
\usepackage{comment}

\usepackage{subfigure}
\usepackage{subcaption}
\usepackage{graphicx}
\usepackage{setspace}
\newcommand{\ts}{\textsuperscript} 

\usepackage{amsmath}
\usepackage{amsthm}
\usepackage{amssymb}
\usepackage{mathrsfs} 
\usepackage{graphicx, color}
\graphicspath{{fig/}}

\titlespacing*{\section}
{0pt}{2.5ex plus 1ex minus .2ex}{1.3ex plus .2ex}
\titlespacing*{\subsection}
{0pt}{1.0ex plus 1ex minus .2ex}{1.3ex plus .2ex}
\titlespacing*{\subsubsection}
{0pt}{1.0ex plus 1ex minus .2ex}{1.3ex plus .2ex}

\title{Temporal Sequencing of Documents}
\author[1]{Michael Gervers}
\author[1]{Gelila Tilahun}
\affil[1]{University of Toronto, Canada}
\corrauthor{Gelila Tilahun}{gelila.tilahun@utoronto.ca}
        
\begin{document}

\maketitle

\abstract{We outline an unsupervised method for temporal rank ordering of sets of historical documents, namely American State of the Union Addresses and DEEDS, a corpus of medieval English property transfer documents. Our method relies upon effectively capturing the gradual change in word usage via a bandwidth estimate for the non-parametric Generalized Linear Models \citep{fan1995local}. The number of possible rank orders needed to search through for cost functions related to the bandwidth can be quite large, even for a small set of documents. We tackle this problem of combinatorial optimization using the Simulated Annealing algorithm, which allows us to obtain the optimal document temporal orders. Our rank ordering method significantly improved the temporal sequencing of both corpora compared to a randomly sequenced baseline. This unsupervised approach should enable the temporal ordering of undated document sets.}  


\section{Introduction}

\label{intro}
The accurate dating of historical and heritage texts is of paramount importance to historians. On the basis of such correctly sequenced texts, historians can examine, judge, and analyze events within the context of a specific time period. Often, only the undated textual contents of the historical documents are available to historians, on the basis of which they must infer the dates of composition \citep{gervers2002dating}. English property-transfer documents (charters or deeds) were selected as one component of this study because of their particular nature. That is, while the earliest surviving examples from the Anglo-Saxon period (c. A.D. 670 to 1066) are invariably dated, only 300 out of a total of approximately 1,600 can be considered originals. Experts have noted that many supposed Anglo-Saxon documents are actually later forgeries, but nonetheless are difficult to distinguish from genuine charters; ``Thus in some cases, the date given is either demonstrably fictional or suspect" (Robert Getz, personal communications, April 25, 2019) or, in many cases, the charters survived only in later copies made centuries after the date of issue, resulting in genuine errors arising from misreading or miscopying by scribes. Common instances include miscopying of Roman numerals, confusion about a witness's identity, or the names of reigning monarchs \citep{whitelock1979Anglo, cubitt1999Catherine}. For examples of documents allegedly forged or fabricated to justify social realities and for political motivations, see \citet{Hiatt2004Alfred}.

When the Anglo-Saxon political and judicial system was largely replaced by that of the Normans following their conquest of England in 1066, an entirely new phenomenon was introduced: the undated charter. From 1066 to circa 1307 (the start of the reign of King Edward II) only about 3\% of the million or more known charters issued bear internal dates. Dating was reintroduced to the royal chancery in 1189 under King Richard the Lionhearted, but the example was not followed by the nobility and commoners for five score years more. Compared to Continental charters, which with few exceptions were regularly dated internally for the duration, the first 600 years of the English charter record has always floated on a sea of incertitude.

In historical research, the one essential principle is to identify the correct order of events. As is evident from the foregoing, that is one of the most difficult tasks in the profession. It is of far greater concern, however, than for historians alone; in fact, it is common to virtually all avenues of scholarship, not to mention of the many institutions upon which literate society depends. Undated documents are everywhere, leaving lawyers, police and spy agencies, forensic linguists, code breakers, artists and art historians, businesses, real estate agents, medical practitioners, military analysts, philosophers (the list is endless), with the responsibility of determining what event preceded or succeeded another. This study sets the stage for anyone with a series of undated, digitized texts, or even lists, to determine a chronological order thereof without having to undertake the arduous task of examining each document for contextual clues and references to specific events, and of identifying periodization through content analysis, handwriting and/or watermarks. All of these aspects can be accomplished automatically through the temporal sequencing methodology outlined below.

By way of examples in which correct temporal ordering was essential, we note that in the financial fraud investigations of Enron Corporation, forensic linguistics was used to analyze emails, memos and internal communications to re-construct the timeline of fraudulent activities even when the timestamps of these evidential materials were not always available \citep{mclean2003smartest}.

The Library of Congress contains many written documents from former presidents of the United States. For example, The Papers of Abraham Lincoln\footnote{\url{https://www.loc.gov/collections/abraham-lincoln-papers/}} is a digitized corpus of over 40K documents consisting of the correspondence, notes, memos, personal letters, drafts of speeches of Abraham Lincoln from his time as a lawyer, congressman and then as the 16\textsuperscript{th} president of the United States. Chronological gaps in The Papers remain, as not all of the original letters and documents were meticulously dated or preserved. A proper chronological order would give insight into the President's evolving thoughts and ideas through a tumultuous period of American History. Another similar example is The Papers of Thomas Jefferson\footnote{\url{https://www.loc.gov/collections/thomas-jefferson-papers/}}, a digitized corpus of 25K items consisting of the correspondence of Thomas Jefferson, who was a diplomat, architect, lawyer and the third president of the United States. Besides correspondence, the collection also includes his drafts of The Declaration of Independence, drafts of laws, fragments of his autobiography, his personal notes, including his records of spending and even recipes! Establishing an accurate chronological order of The Papers is crucial in understanding the personal worldview and the evolving visions about the early Republic by one of the prominent Founding Fathers.

The medieval Exeter Book\footnote{\url{https://www.exeter-cathedral.org.uk/learning-collections/explore-the-collections/the-exeter-book/}} is another example. It is an anthology of Old English poetry and riddles from the 10\textsuperscript{th} century, but the chronological order of none of the texts is known. Establishing a chronological order of the texts would give us a deeper understanding of the evolution of Old English and the literary culture of the Anglo-Saxon people.

Previous efforts in document sorting have been directed towards the development of historical language models \citep{feuerverger2005distance, feuerverger2008using, tilahun2016statistical, gervers2018thedating}. Within the  broader field of information retrieval, investigators employ statistical models incorporating temporal aspects of term usage \citep{swan2000timemines}, study the relationship between time and the retrieval of relevant documents \citep{li2003time}, and classify document dates according to time partitions of predefined granularity \citep{de2005temporal}. In preparing web pages that present document lists the accuracy of time-stamping is of paramount importance. In this regard, \citet{kanhabua2008improving} and \citet{kanhabua2009using} extended \citet{de2005temporal}'s work by integrating semantic-based techniques into the document pre-processing pipeline, the aim being to improve the temporal precision of web pages and web document searches where the trustworthiness of document time-stamps is often questionable. \citet{lebanon2008local} modelled temporal text streams, and \citet{mani2000robust} extracted temporal expressions such as {\it now}, {\it today}, and {\it tomorrow} from print and broadcast news documents to resolve uncertain temporal expressions. \citet{chambers2012labeling} presented models that incorporate rich linguistic features about time, whereas  \citet{vashishth2019dating} employed deep neural network-based methods that also exploit linguistic features about time in order to date the documents. However, as pointed out by \citet{kotsakos2014burstiness}, relying upon temporal terms for dating suffers from the drawback that terms ``can be very sparse as well as ambiguous, referring to irrelevant timeframes". These authors proposed using statistical approaches based on lexical similarity and burstiness --- the sudden increase in the frequency of terms within a time-frame. 

In the current work, we propose TempSeq\footnote{The TempSeq pseudo and source codes used in this paper are available at \url{https://github.com/gitgelila/TempSeq} }, an unsupervised method for the temporal \emph{sequencing} or \emph{ranking} of documents. This approach is designed to be applicable when the only available data are the undated documents that are to be temporally sequenced. TempSeq relies on a `bag-of-words' approach, and does not make use of linguistic features about time, nor does it use a training set data with time-tag. In addition, our approach does not make use of specific language rules, word representations, or any other metadata information, thus presenting a potentially significant advantage in the task of document temporal ordering. TempSeq relies on measuring word usage drift under the assumption that word usage changes gradually over time, which means that the temporal variability in word usage is low. We model word usage drift via the non-parametric Generalized Linear Models regression \citep{fan1995local}, and we estimate the correct temporal sequencing of the documents to be the one that minimizes, on average, the associated kernel bandwidths (a direct measure of the temporal variability in word usage). To our knowledge, using the variability of word usage drift to ascribe a temporal sequencing for a set of documents is entirely new. 

The necessity for temporal sequencing of documents arises not only in the field of information retrieval, but also in studies of heritage texts, which frequently lack timestamps, are intentionally ambiguous with respect to time of issue, or can even be outright forgeries. Often, only the textual contents of heritage documents are available, from which one must infer the dates of issue \citep{gervers2002dating}. Furthermore, heritage texts that are available as a training data set can be limited in number, as proportionately few documents have survived across the centuries to the present time, thus necessitating an unsupervised method for inferring document dates. The task at hand is not only to infer the temporal ranking of a collection of documents or corpus, but also to identify the terms that contribute most towards the task of identifying the correct temporal ranking/ordering. We believe that such terms are most likely the temporal signatures for identifying the characteristics of intentionally or inadvertently temporally mislabelled documents, or documents with missing or corrupted timestamps.     

Past and present research about problems arising in document sequencing involve criteria for document classification, for example, by topic,  \citet{blei2003latent,blei2006dynamic, mcauliffe2008supervised, taddy2013multinomial,taddy2015document}, document indexing \citet{roberts2016model}, and document ranking using latent semantic analysis \cite{deerwester1990indexing, hofmann1999probabilistic}.  \citet{cohen1998learning} consider the problem of machine learning to order instances, as opposed to classifying them, when an algorithm's output receives feedback in the form of \emph{preference judgments}, i.e., ``statements indicating which instances should be ranked ahead of the others" \citep{cohen1998learning}. However, such lines of research have not directly addressed the problem of document ordering {\it per se}. Perhaps the closest approach to dealing with this problem is that of Thinninyam in her dissertation \citep{thinniyam2014statistical}. Her approach is based on the notion of similarity (or distance) between two documents, namely the supposition that similar documents are more likely to discuss similar topics, and therefore should have closer underlying temporal footprints. She proposed a linear regression-based approach, with regression of observed document distance measures of spacings between consecutive documents. In a separate approach, Thinnayam also framed the document ordering problem in a Bayesian framework, where the distribution of pairwise distances between documents was modelled conditionally on a timeline vector, where the $i^{th}$ coordinate value of the vector represents the time interval between the $i^{th}$ document and a reference document. Herein, a Markov Chain Monte Carlo (MCMC) method was employed to sample from the posterior distribution of the timeline. Thinniyam's ordering methods fundamentally require pairwise distance measures of documents (i.e., a quantifiable measure of dissimilarity between two documents) to estimate the temporal order of a set of documents within a corpus. Other studies have shown that such measures are prone to yield spuriously high values of similarity due to an abundance of uninformative terms within the documents, including, but not limited only to stop words \citep{tilahun2011statistical}; it is not always a straightforward matter to identify these uninformative terms, and the requisite degree of filtration.  Moreover, using measures of document distance does not allow identification of the particular words or terms that are essential for determining the predicted temporal rank orders. In addition, the degree to which two documents are similar/dissimilar is highly dependent on the type of distance measures that are used \citet{broder1997resemblance}.

In contrast to Thinnyam's previous approach, the present TempSeq method temporally ranks a set of documents even when a reliably dated training dataset is not available, and/or when there is a very limited number of documents in the set. The TempSeq method relies fundamentally on modelling the probability of occurrence of words in a given date range, thereby avoiding the need for a document distance measure. By design, TempSeq also filters out words according to their degree of uninformativeness, thereby allowing us to gain insights into the history underlying the documents by identifying words that are most putatively useful for determining the correct temporal ordering of documents. We test the TempSeq method on two corpora of heritage texts; one written in American English and the other in Latin. When a set of training data is available, TempSeq can optimize the smoothing parameter for temporal sequencing. More importantly, we show for both corpora that the TempSeq temporal sequencing method performed significantly better as compared to random sequencing, in the absence of training data.

\section{Corpora}
\label{corpus}
We evaluated our temporal sequencing methods on two different sets of corpora with time-tags. The first corpus consisted of 240 transcripts of the American State of the Union Address (SOTU), from the years 1790 to 2020. Each transcript had a median average length of 6400 words. This corpus is available from the R package, \citet{Taylor2022}. The second corpus is from the Documents of Early England Data Set (DEEDS)\footnote{\url{https://deeds.library.utoronto.ca}}. From within this corpus, we focused on a set of 11,463 English property conveyance records issued in the years 1120 to 1300. All the records are written in Latin, and have been inspected for content by subject expert historians to accurately verify the date of issue. The Latin documents have a median length of 175 words. We chose to evaluate the TempSeq temporal document sequencing method on this corpus as it consists of documents similar to corresponding DEEDS documents from the Anglo-Saxon period, which, as mentioned in section~\ref{intro}, have generally unreliable dates. In this project, we considered the DEEDS corpus in two different forms. In the first form, we conflate all the documents written in a given year into single texts, thus yielding 181 conflated DEEDS records, of mean average length of approximately 11,000 words. We denote the conflated collection as ``DEEDS-conflated". In the second form, we denote the entire set of 11,463 unconflated records as ``DEEDS-single".

\section{Outline}
When a set of training documents with known dates  
is available, \citet{tilahun2012dating} have proposed 
the ``Maximum Prevalence" method for their dating. This approach is based on modelling (on the basis of the training data) a curve that describes the temporal pattern of the probability of occurrence of each word from the undated documents. For example, in the DEEDS corpus, the proposed dating method achieves very reliable date estimates, giving a test set median dating error of $\pm$5 years within the 230 year span ~\citep{gervers2018thedating}. High accuracy validates an underlying feature of the model, namely useful words for dating a document are those with a non-uniform probability of occurrence across a date range, and showing a gradual change in the variability of their usage changes. Words such as {\it et, de, huic} (in Latin), {\it the, to, that, on} (in English), and stop words, which appear in consistent proportion at all times, that is to say {\it uninformative} words, do not contribute to the date estimation of an undated document.

In section~\ref{theonew}, we discuss our modelling approach to estimate the curves best describing the temporal pattern of the probability of occurrence of a given word/phrase, and present examples of such curves. 
We also examine the properties of the curve estimates (in particular, that of a smoothing parameter) in relation to the bias-variance trade-off, using the form of the bias-variance trade-off to select the optimal curve. This trade-off is at the heart of any statistical learning process. We could perfectly fit training data (zero bias) to a model by including excessive amounts of parameters (excessive under-smoothing in the case of our model). This situation, referred to as over-parametrization, risks overfitting the data because the model learns not only the pattern in the data but also the random noise and fluctuations that are present. When an overfitted model is applied to a test data, the performance is often poor. On the other extreme, when a model is under-fitted (over-smoothed in the case of our model) the effect from random noise is eliminated but at the expense of failing to learn the pattern from the data. The right amount of parameterization is one that balances bias and variance (large bias and low variance). We seek an optimization process which can balance this trade-off. This optimization seeks on the one hand to minimize bias, thereby increasing curve fluctuation to accurately track the empirical values of the probability of word occurrences. At the same time, the optimization minimizes variance, thereby decreasing the amount of curve fluctuation to obtain a smooth curve. The optimally smoothed curve for balancing the trade-off between these demands is a quantifiable parameter value that can be estimated using a ``rule-of-thumb" smoothing parameter estimate \citep{fan1995local}. In section~\ref{optimize}, we address the problem of temporally ordering a set of documents in the absence of a set of dated training data. To this end, we compute the average value of the optimal smoothing parameters for estimating the probability of occurrence of each word in the documents. Here, we find a close estimate of the correct temporal order of well-spaced subsets of ($m=10$) documents by searching among all possible temporal orderings to identify the highest average optimal smoothing parameter.  We carried out this search using combinatorial optimization via the Simulated Annealing algorithm. In section~\ref{results}, we evaluate the TempSeq method and present its results for the two distinct corpora --- the DEEDS corpus and the SOTU corpus. In addition, we identify the informative words that enabled TempSeq to establish the correct temporal order for the selected subset of documents.  In section \ref{error}, we provide error analysis, and present our general conclusions in section~\ref{discussion}. Theoretical background and operational equations are presented in Annex~\ref{appendix} and Annex~\ref{appendix2}. 

\section{Modelling the Temporal Pattern of Word Usage}
\label{theonew}
Our fundamental assumption is that word usage changes gradually.
We model the probability of word usage as a function of time using the local polynomial kernel regression for the generalized linear model,~\cite{fan1995local}. For further details of this model, see Annex~\ref{appendix}, section~\ref{appendix1}. 

Suppose $(D_{i}, t_{D_{i}}), i=1, \ldots, n$ represents a sequence of data pairs, where $t_{D_{i}}$ represents the date of the ${D_{i}}^{{th}}$ document and $n$ denotes the size of the data set, that is to say, the total number of documents in the collection.  Let {$n_{w}(D_{i})$} denote the number of occurrences of the word (or term) $w$ in document $D_{i}$. Finally, let $N(D_{i})$ denote the total number of words (or terms) in document $D_{i}$. We are interested in estimating the probability of occurrence of the term {\em w} at time $t$, which is given by:
\begin{eqnarray}
\label{prob}
	 \hat{\pi}_{w,h}(t)=\frac{\sum_{i=1}^{n}n_{w}(D_{i})K_{h}(t_{D_{i}}-t)}{\sum_{i=1}^{n}N(D_{i})K_{h}(t_{D_{i}}-t)}.
\end{eqnarray}

We define the weight term $K_{h}$ to be $K_{h}(u)\equiv h^{-1}K(u/h)$, where $K$ is called a kernel function. Typically, $K$ is a bell-shaped, non-negative function, with an area under the graph equalling unity. 
The function decays fast enough to eliminate the contributions of remote data points. (See equation~\ref{probsp0}, in Annex~\ref{appendix}, section~\ref{appendix1}). For the present study, we used the t-distribution function with a low degree of freedom value (equal to $5$) to adequately weigh distant points. 

As a weight term, $K_{h}$ fits a polynomial regression curve around the data in the neighbourhood of $t$, where $h$, called the bandwidth parameter, is the size of the local neighbourhood, such that data points distal from the neighbourhood are down-weighed (for this reason, $h$ is also referred to as the curve smoothing parameter). In simple terms, if $h$ is very large (highly smoothed), then $\hat{\pi}_{w,h}(t) \approx \sum_{i=1}^{n}n_{w}(D_{i})/\sum_{i=1}^{n}N(D_{i})$, thus representing an overall proportional outcome of word $w$ which does not change with $t$. On the other extreme, if $h$ is very small, then, $\hat{\pi}_{w,h}(t)$ evaluated at, say, $t_{D_{j}}$, has the value $\hat{\pi}_{w,h}(t_{D_{j}}) \approx  n_{w}(D_{j})/N(D_{j})$, which is the proportional outcome of word $w$ in the document written at time $t_{D_{j}}$. In this case, information on the frequency of occurrence of word $w$ in documents written at dates near to $t_{D_{j}}$ has been completely ignored in determining the value of $\hat{\pi}_{w,h}(t_{D_{j}})$. When $h$ is very small, the curve $\hat{\pi}_{w,h}(t)$ overfits, thus fluctuating rapidly to attain the values $n_{w}(D_{i})/N(D_{i})$ for each time point $t_{D_{i}}, i=1, \ldots, n$. Although it is possible to draw a curve that perfectly describes the empirical probability of occurrence of a word across a date range (i.e., a bias with a value of zero), the consequent high variance of the curve means that, when applied to a test data set, the curve would overfit thus depicting a very inaccurate description of the probability of occurrence of the word of interest. In the field of kernel regression, there has been extensive research on how to select the appropriate bandwidth parameter $h$. However, in implementing the TempSeq method, we have relied on using a rule-of-thumb selection approach (\cite{fan1995local}), which we describe in Annex~\ref{appendix2}, section~\ref{appendix2a}. The theoretical details for the derivation of  equation~\ref{prob}, which falls under the \textit{locally constant case} ($p=0$), can be found in Annex~\ref{appendix}, section~\ref{appendix1}.

\begin{figure}[!htb]
\begin{center}
\includegraphics[width=10cm,height=8cm]{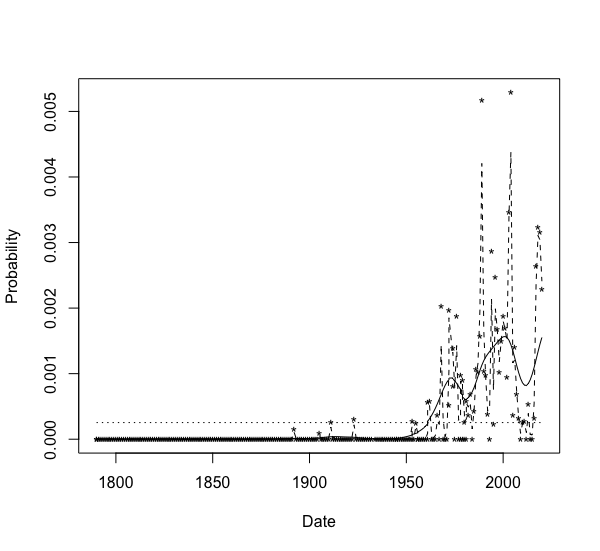}
\caption{Asterisks show the proportion of occurrences of the words {\it Drug(s)} in the SOTU corpus. The solid curve is based on a larger bandwidth value than that of the dashed-lined curve. The dotted curve (the horizontal dotted line) is based on a very large bandwidth value. Date (time) is the $x$-axis and $\hat{\pi}_{w,h}(t)$ is the $y$-axis.}
\label{Drug}
\end{center}
\end{figure}

Relying on our assumption that change in word usage is gradual, we now focus on the role played by the bandwidth parameter $h$ in setting the bias-variance trade-off of the estimator $\hat{\pi}_{w,h}(t)$. Figure~\ref{Drug} illustrates the probability of occurrence of the words {\it Drug(s)} in the SOTU corpus at different dates and for variable bandwidth settings. The $x$-axis illustrates the calendar year (the time {\it t}), ranging from 1790 to 2020. The $y$-axis illustrates the values of $\hat{\pi}_{w,h}(t)$. The asterisks show the proportion of occurrences of the words ``Drug(s)" in the years for which dated SOTU documents are available. The proportion is greater than zero in a few of the years, but is zero for a few years preceding or following that time. In that circumstance, when the bandwidth value {\it h} is small, the resulting estimate, $\hat{\pi}_{w,h}(t)$, is highly variable (the closer {\it h} approaches zero, the closer its values match the recorded proportion of occurrence of the term in the training data). This behaviour is illustrated by the dashed-line curve of $\hat{\pi}_{w,h}(t)$ in the figure. When the value of $h$ is larger (smoother), then the resultant probability curve $\hat{\pi}_{w,h}(t)$, overlaid in solid, has less variability.  Conversely, when the value of $h$ is very large, the closer the values of $\hat{\pi}_{w,h}(t)$, for all $t$, approach the proportion of the real occurrence of the term $w$ across all the dates in the document set (illustrated in dotted lines). The bandwidth, which controls both the bias and the variance, is therefore a crucial parameter of the estimator $\hat{\pi}_{w,h}(t)$. 

The optimal amount of smoothing of the data in figure~\ref{Drug} results in the solid curve, which illuminates a clear pattern in the data. The first peak in the figure (coinciding with the presidency of Richard M. Nixon (1970-1974)) refers to the emergence of the so-called {\it War on Drugs}, a US-led global policy aimed at the production, distribution, and use of psychoactive drugs, which was presented in Nixon's State of the Union address of 1972. That address contains the phrases `{\it ... strong new drug treatment programs ...}', `{\it ... by continuing our intensified war on drug abuse ...}', `{\it ... Special Action Office for Drug Abuse Prevention ...}', `{\it ... collective effort by nations throughout the world to eliminate drugs at their source ...}', `{\it ... to drive drug traffickers and pushers off the streets of America ...}', `{\it ... to curb illicit drug traffic at our borders and within our country ...}'. His 1974 address contains the phrases `{\it ... the spiraling rise in drug addiction ... }', `{\it ... The Psychotropic Convention ... treaty regulating manufactured drugs worldwide ...}' and `{\it ... the drug battle is far from over ...}'.  The first peak extends to President Gerald Ford's 1976 address, which contains phrases such as, `{\it The sale of hard drugs is tragically on the increase again ...}' and `{\it ... shipment of hard drugs ...}'. The second peak in figure~\ref{Drug} occurs decades later, around the year 2000. Then President William Clinton's 1998 address contains phrases such as, `{\it ... to crack down on gangs and guns and drugs ...}' and `{\it ... the largest anti-drug budget in history ...}';  his 1999 address has phrases such as  `{\it ... if you stay on drugs, you have to stay behind bars ...}' and `{\it ... to strengthen the Safe and Drug-Free School Act ...}'. In his 2000 address, Clinton announced `{\it ... new legislation to go after what these drug barons value the most, their money.}'. Clinton also invokes the words {\it drug(s)}, in the positive sense of insurance coverage for affordable prescription drugs. In that context, he used phrases such as `{\it ... seniors now lack dependable drug coverage ...}' and `{\it Lifesaving drugs are an indispensable part of modern medicine ...}'. In the following years, under President George W. Bush, policy regarding affordable drug coverage becomes a major issue domestically and globally; `{\it ... some form of prescription drug coverage ...}' (in the 2001 address), `{\it ... new drugs that are transforming health care in America ...}' (in the 2003 address) and `{\it More than 4 million require immediate drug treatment}' (in the 2003 address regarding the lack of antiretroviral drugs in Africa). Thus, the first peak in figure \ref{Drug} is exclusively related to illicit drug issues, whereas the second peak, some 25 years later, is primarily related to the affordability of prescription drugs.

\begin{figure}[!htb]
\begin{center}
\includegraphics[width=10cm,height=8cm]{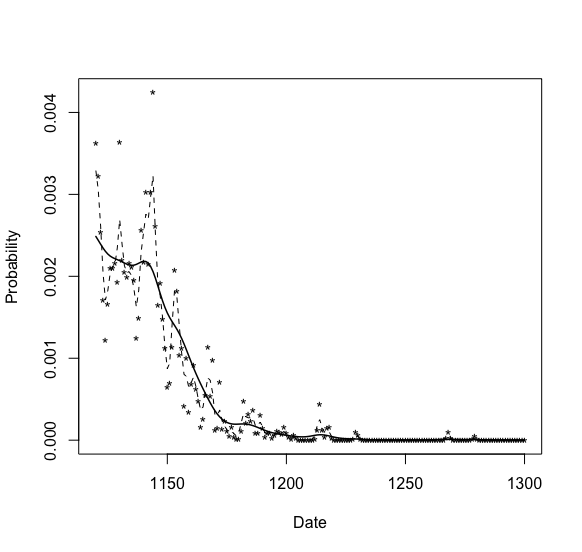}
\caption{Asterisks show the proportion of occurrences of the phrase {\it Angl(ic)is} in the DEEDS corpus. The solid curve is based on a larger bandwidth value than that of the dashed-lined curve. Date (time) is the $x$-axis and $\hat{\pi}_{w,h}(t)$ is the $y$-axis.}
\label{Anglis}
\end{center}
\end{figure}

Figure~\ref{Anglis} illustrates the probability of occurrence of the words {\it Anglicis} and {\it Anglis} in the DEEDS corpus. These words are often found within the form of address `{\it Franc[igen]is quam(et) Angl[ic]is}' (French and English), such as `{\it ... tam presentibus quam futuris tam Francigen[is] quam Anglicis salutem Sciatis me intuitu dei assensu ...}', (... both present and future, both French and English, greeting. Know that with God's consent I have [granted] ...) and  `{\it ... omnibus ministris et fidelibus suis Francis et Anglis de Oxenfordscira ...}',  ({... to all his French and English ministers and servants of Oxfordshire ...}). The above form of address was commonly used by French and the English barons of the time to address their subjects. However, after the province of Normandy was conquered by the French in 1204, this form of address virtually fell out of use.  

\begin{figure}[!htb]
\begin{center}
\includegraphics[width=10cm,height=6.5cm]{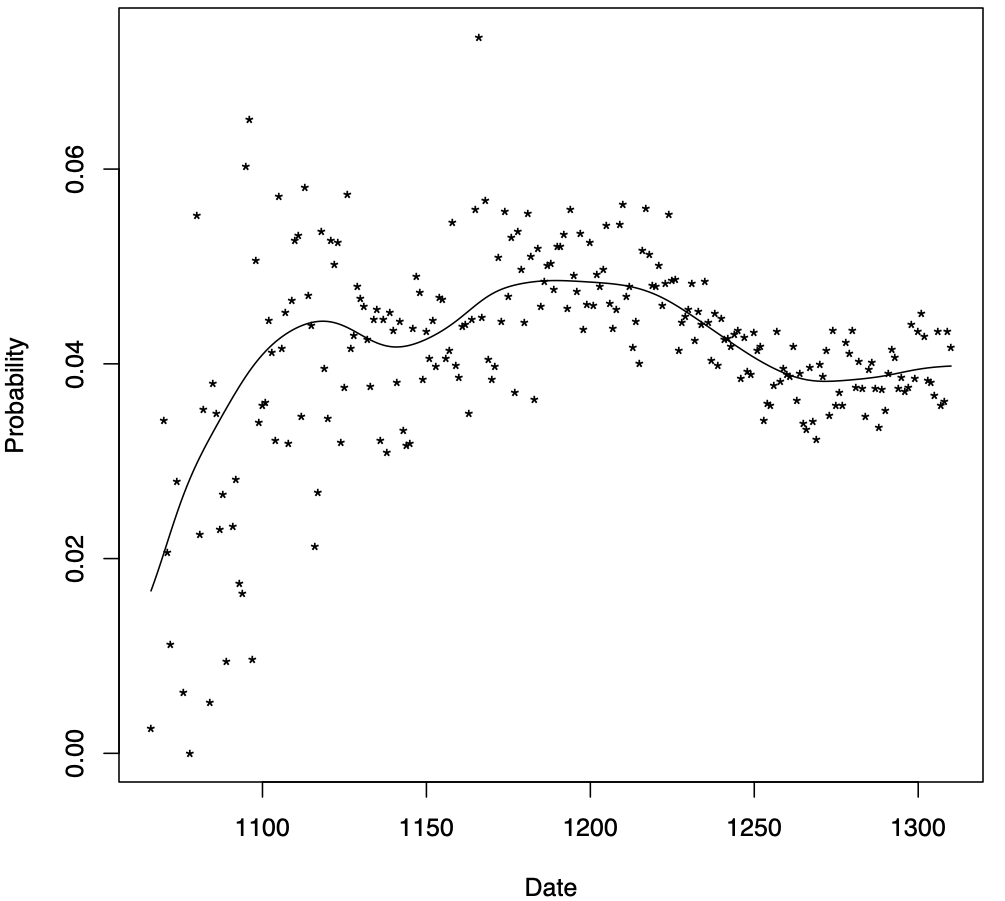}
\caption{Asterisks show the proportion of occurrences of the word {\it de} (of). The smoothed solid probability curve is uniform across the date range. Date (time) is the $x$-axis and $\hat{\pi}_{w,h}(t)$ is the $y$-axis.}
\label{De}
\end{center}
\end{figure}

Figure~\ref{De} illustrates the probability of occurrence of a common stop word `{\it de}' (of) in the DEEDS corpus. The asterisks in the figure show the proportion of occurrences of the word across time (1120 to 1300), and the line curve is found by smoothing the proportion of occurrences of the word across those same years. When comparing the smoothed black curves in figure \ref{Drug} to figure \ref{De}, we see that the latter curve is more uniform across time (except for the years prior to 1125, when fewer documents were available). This behaviour, which we call the principle of temporal uniformity (non-uniformity) of uninformative (informative) words, is shown in the figure where the probability of occurrence of {\it de} has no defining temporal feature, and is uniform across the date range. For the purpose of temporally ordering a set of documents, no matter what combination of temporal ordering is evaluated in the TempSeq process, the contribution of uninformative words is immaterial.   
 
In all of these figures \ref{Drug} to \ref{De}, we see that the solid black curves have relatively the most optimal smoothing, as opposed, for example, to the highly variable dashed-line curve in figure \ref{Drug}. The optimal smoothing for a given curve represents the trade-off between small bias and small variance for the curve estimator. If the data analyst then randomizes the true temporal ordering of word usage, applying the optimal smoothing parameter will now result in a curve estimate that rapidly oscillates (high variance) due to seeking a minimum bias.    

\clearpage

\section{The TempSeq Method for Temporal Sequencing}
\label{optimize}
\subsection{Determining the Optimal Bandwidth}
Let $\{D_{1}, D_{2}, \ldots, D_{m}\}$ be a set of $m$ number of documents that we wish to sequence in temporal order. We assume that all the documents have a unique timestamp, and as such, there are $m!/2$ possible orders when two orderings that are the reverse of each other are equated. Without loss of generality, assume that the sequence $l=(1,2, \ldots, m)$ represents the true temporal rank order of the $m$ documents. Let $\sigma(l)$ represent a permutation of the ranks and let  $\sigma_{0}$ represent permutation identity, that is to say $\sigma_{0}(l)=l$. For each word $w$, $w \in \cup_{i=1}^{m} D_{i}$, and temporal rank ordering of the documents under $\sigma(l)$), we compute the asymptotically optimal bandwidth value for $\hat{\pi}_{w,h}(t)$, which we denote as $h_{amise, w,\sigma(l)}$. This bandwidth value is estimated via a `rule-of-thumb' estimate, the detail of which can be found in Annex~\ref{appendix2}, sections~\ref{appendix2a} and~\ref{appendix2b}. 
The formulation of $\hat{\pi}_{w,h}(t)$ here is subject to the condition $p=1$ (the case of the locally linear regression, equation~\ref{probsp1}), which is more accurate than the formulation of $\hat{\pi}_{w,h}(t)$ in equation~\ref{prob} (the case of locally constant regression, $p=0$). For theoretical details, refer to Annex~\ref{appendix}, section~\ref{appendix1}. 

Following the principle of temporal non-uniformity of informative words, the optimal smoothing parameter $h_{amise, w,\sigma(l)}$ will be larger under the correct temporal ordering of the documents, since the curve would not entail such extensive oscillation to obtain a small bias. Therefore, we would generally expect 
\begin{eqnarray}
\label{h0>hl}
	    h_{amise, w,\sigma_{0}(l)}\geq h_{amise, w,\sigma(l)}
\end{eqnarray}
to hold for each word $w$. Put another way, the rule-of-thumb bandwidth estimate of a word associated with the correct temporal ordering of documents will be larger than those bandwidth estimates based on incorrect temporal orderings. For a set of documents $\{D_{1}, D_{2}, \ldots, D_{m}\}$, we estimate the temporal rank order for the set of documents by first computing 
\begin{eqnarray}
\label{sigmahatFirst}
 H_{\sigma(l)}=\underset{w}{\mbox{Median}}\left\{h_{amise, w,\sigma(l)}:  w \in \cup_{i=1}^{m} D_{i}\right\} 
\end{eqnarray}
where $H_{\sigma(l)}$ is the uniform median value of the optimal bandwidths associated with each word\footnote{ For a word $w$ to be included in the estimations, we required that it occurs in at least two documents as measuring the pattern of word usage fluctuations is a key element.} present in the $m$ number of documents, and $\sigma(l)$ is a proposed temporal ordering. The temporal rank order estimate, $\hat{\sigma}(l)$, is the rank order which maximizes the term $H_{\sigma(l)}$ over all possible permutations; stated more succinctly,
\begin{eqnarray}
\label{sigmahat}
     \hat{\sigma}(l) = \mbox{arg}\max_{\sigma} H_{\sigma(l)}.
\end{eqnarray}
The estimated temporal rank order is one that results, on average, in the smoothest rule-of-thumb bandwidth estimate of $h_{amise, w,\sigma(l)}$ over all the words in the $m$ number of documents and over all possible temporal rank orders.

\begin{figure}
\label{FigRandomComp}
    \hfill
    \subfigure[The State of the Union Address]{\includegraphics[width=7cm,height=8cm]{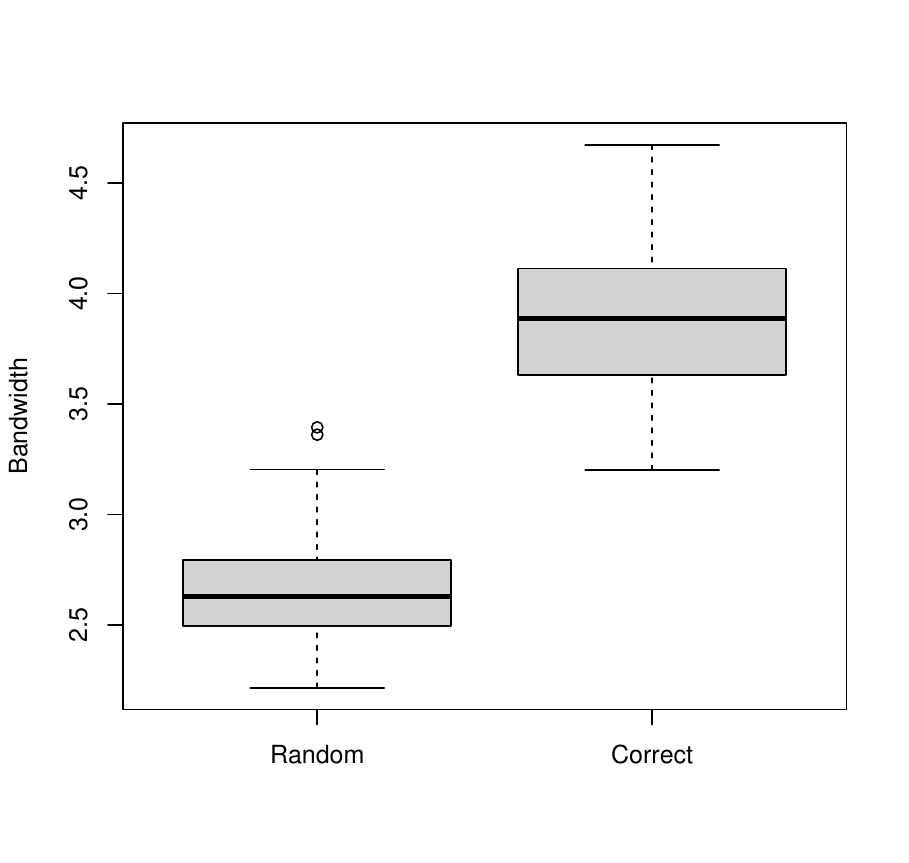} \label{boxplotSofUHsigma}}
    \hfill
    \subfigure[DEEDS-conflated]{\includegraphics[width=7cm,height=8cm]{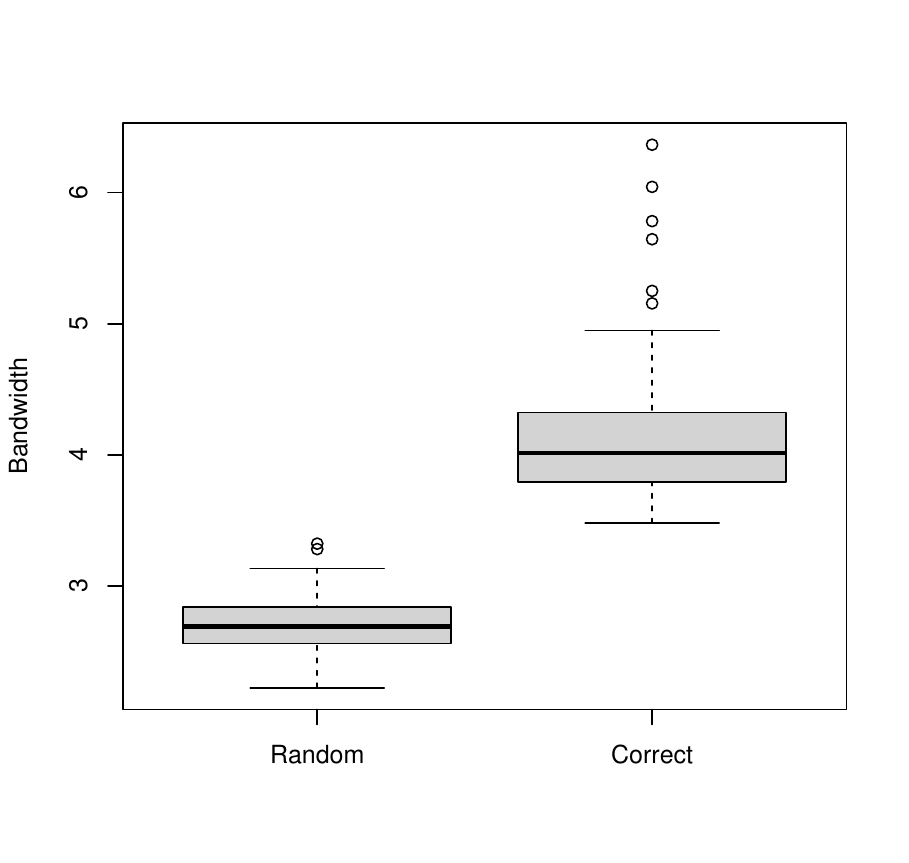} \label{boxplotDEEDSHsigma}}
    \centering
    \subfigure[DEEDS-single]{\includegraphics[width=7cm,height=8cm]{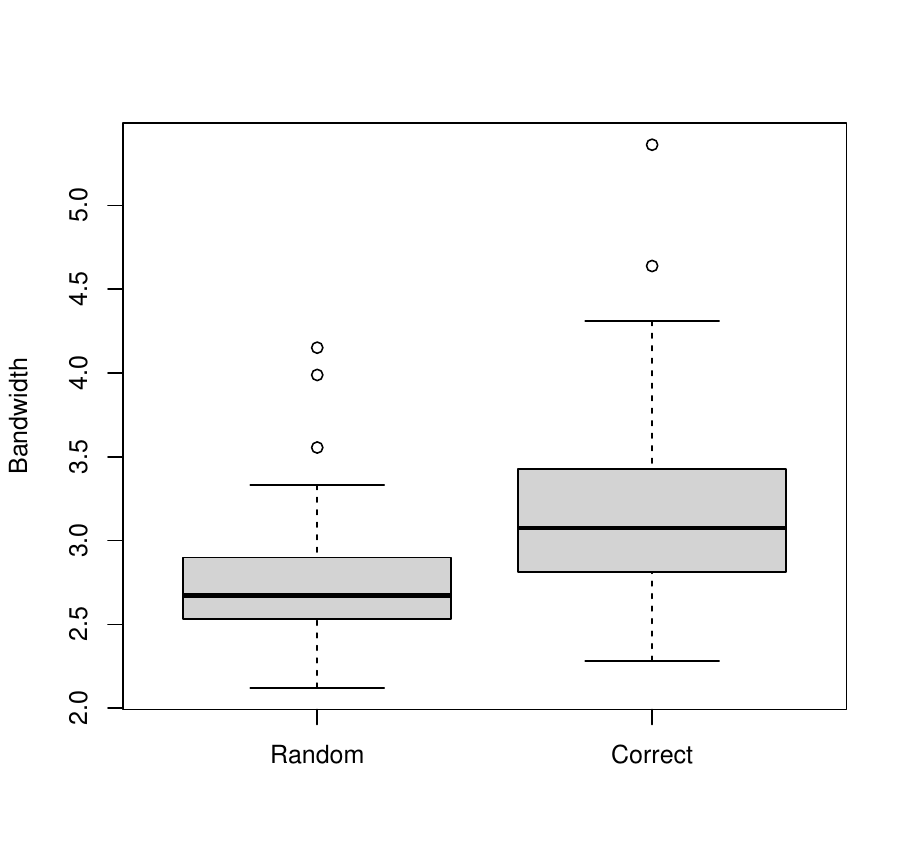} \label{boxplotDEEDSSingleHsigma}}
    \caption{Box plots of $H_{\sigma(l)}$ (Bandwidths) versus temporal orders for 100 randomly selected sets of ten documents.  In all figures, the first two box plots are that of $H_{\sigma(l)}$ for temporally randomly permuted and unpermuted sets of documents. Figure \ref{boxplotSofUHsigma} shows the results for the SOTU corpus. Figures \ref{boxplotDEEDSHsigma} and \ref{boxplotDEEDSSingleHsigma} show the results for the DEEDS corpus (DEEDS-conflated and DEEDS-single, respectively).}
\label{Hsigmas}
\end{figure}

To verify our expectation that equation (\ref{h0>hl}) in fact holds in general, which would imply that $H_{\sigma_{0}(l)} \geq H_{\sigma(l)}$ also holds, we conducted the following experiment separately on the DEEDS and the SOTU corpora. In each case, we randomly selected a set of ten documents with date gaps of about 20 years, i.e., one tenth of the document history, thus obtaining a trade-off between excessive computational time and fitness of the method for correct ordering. In this computation, only those words that occurred at least once in two separate documents were considered. Based on random permutations of the underlying true temporal rank order, $\sigma(l)$, we computed $H_{\sigma(l)}$. For the same set of ten documents, we also computed $H_{\sigma_{0}(l)}$, when the true temporal order of the documents was not permuted. We ran 100 replications of the above experiment. Figures \ref{boxplotSofUHsigma}, \ref{boxplotDEEDSHsigma} and \ref{boxplotDEEDSSingleHsigma} are the box plots of $H_{\sigma(l)}$ (bandwidths) for the SOTU, DEEDS-conflated and DEEDS-single corpora, respectively. In each figure, the first box plot is that of $H_{\sigma(l)}$ where $\sigma(l)$ is a random permutation of the true temporal order of the given set of ten documents, and the second box plot depicts the case when the true temporal order of the same ten documents is maintained. As shown by these box plots, optimal bandwidths associated with the true temporal orderings are generally larger (smoother) than those associated with random orderings. In comparing all the box plots in the figures, we see that those associated with DEEDS-single more closely resemble one another. This result is not surprising, since the computation of equation~\ref{sigmahatFirst} on sets of documents relies upon fewer words than those from the SOTU and DEEDS-conflated corpora. 

We note that no matter what the permutation of the underlying documents' sequence, the bandwidth $h_{amise}$ associated with uninformative words remains unchanged. Therefore, the contribution of uninformative words has negligible influence on the estimation of the temporal rank order of the set of documents.

\subsection{Estimation via Simulated Annealing}
With an increasing number of documents that we wish to place in order, there is a corresponding increase in the number of permutations required to search exhaustively in order to obtain $\hat{\sigma}(l)$ (equation \ref{sigmahat}). For example, when $m=10$, the requisite number of permutations equals ${10!}/{2}$ (circa 1.8 million), where a permutation and its reverse are equated. Scaling up to a large number of possible permutations to optimize in an objective function (such as equation \ref{sigmahat}) calls for combinatorial optimization. We propose to solve this problem using the well-known Simulated Annealing algorithm \citep{kirkpatrick1983optimization}. 
	
In the current task, we are attempting to find permutations of the temporal rank order of the document sets that maximize $H_{\sigma(l)}$.  The optimization problem involves a search over the neighbours of a permutation element $\sigma(l)$, and the generating scheme of the candidate solution (neighbourhood) along with its set size are important factors in the performance of Simulated Annealing \citep{tian1999application}. We employ a neighbourhood generating scheme proposed by those authors for the well known Travelling Salesman Problem. The proposed scheme generates a random permutation solution from the current one by reversing and/or moving a subsequence of terms. For example, the sequence $\{1, 2, 3, 4, \underline{5, 6, 7, 8}, 9,10\}$, could generate the $\{1, \underline{8, 7, 6, 5}, 2, 3, 4, 9,10\}$. In fact, this perturbation scheme (where a random set of subsequences with four terms that are randomly reversed and/or moved), was employed in this paper to generate the candidate neighbours for the Simulated Annealing algorithm. The authors prove that under such a random perturbation scheme to generate random permutation solutions, the Simulated Annealing algorithm converges asymptotically to the set of global optimal solutions.

\section{Evaluation and Results}
\label{results}
For a task of ordering temporally a set of $m$ number of documents ${\{D_{1}, D_{2}, \ldots, D_{m}\}}$, as in section \ref{optimize}, let the true temporal rank order of the documents be ${ l=(1,2, \ldots, m)}$. Let $\sigma(l)$ represent a permutation of the ranks and let $\sigma_{0}$ represent the permutation identity ($\sigma_{0}(l)=l$). We measure the extent to which two permutations are in close proximity to one another using the Spearman's ($\rho$) rank correlation. If $\hat{\sigma}(l)$ is the predicted temporal rank order for the set of documents ${ \{D_{1}, D_{2}, \ldots, D_{m}\}}$, then the closer ${ | \rho(\hat{\sigma}(l), \sigma_{0}(l)) |}$ is to unity, the more accurately the predicted order matches the true order (we only consider the absolute value of the correlation since forward and reverse orders are equated, as noted above).

\begin{figure}[t]
\begin{center}
\includegraphics[ width=10.0cm,height=7cm]{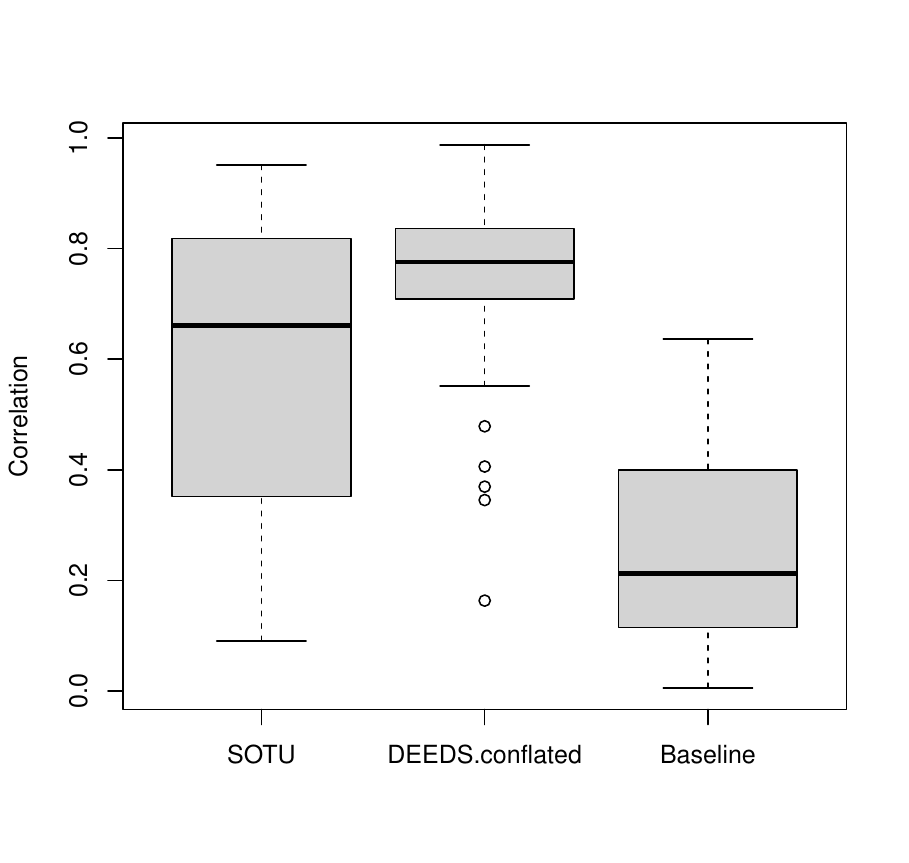}
\caption{Box plots of the correlation coefficients (in absolute terms) of the estimated rank orders of sets of 10 documents and their true rank orders, replicated 100 times. The first plot corresponds to the State of the Union Address corpus (SOTU), the second to the DEEDS-conflated corpus, and the final plot is the baseline (random).}
\label{boxplotComp}
\end{center}
\end{figure}

We randomly selected sets of 10 documents, dated approximately 24 years apart for the SOTU, and 18 years apart for the DEEDS-conflate corpora. For the random selection, we used systematic sampling, as follows: First, we randomly selected the start date document, and then every 24\textsuperscript{th} year in succession for the annual SOTU, and then for ever 18\textsuperscript{th} year for the annual conflation of DEEDS documents (DEEDS-conflated), or the corresponding randomly selected single DEEDS documents (DEEDS-single). If the subsequently selected year exceeded the range of dates, we cycled through from the start date. Then, we labelled the temporal rank of the resulting documents from 1 to 10. Starting from an initial random temporal order of the ten documents, we estimated their true temporal rank order as described above, and calculated the Spearman correlation for the true and shuffled ordering. In all cases, our analysis was based on words that occurred at least once in two separate documents (otherwise, no information regarding temporal ordering can be inferred). For 100 replications of the above procedure, the median of the absolute value of the correlations between the estimated and the true rank orders was 0.66 for the SOTU corpus and 0.78 for the annually conflated DEEDS corpus. As a baseline comparison, we note that the median of the absolute value of the correlations between the true temporal rank orders and their 100 random permutations was 0.24. In the computation, we only considered words that occurred more than once in the set of 10 documents. As shown in the box plots of figure \ref{boxplotComp}, the TempSeq method performed significantly better than the baseline correlation. The Wilcox rank sum test and the t-test showed a statistically significant difference between the baseline correlation and the correlations associated with each of the SOTU and the DEEDS-conflated corpora. In both cases, $\mbox{p-value} < 2.2\times 10^{-16}$. 
	
\begin{figure}[!htb]
\begin{center}
\includegraphics[ width=10.0cm,height=7cm]{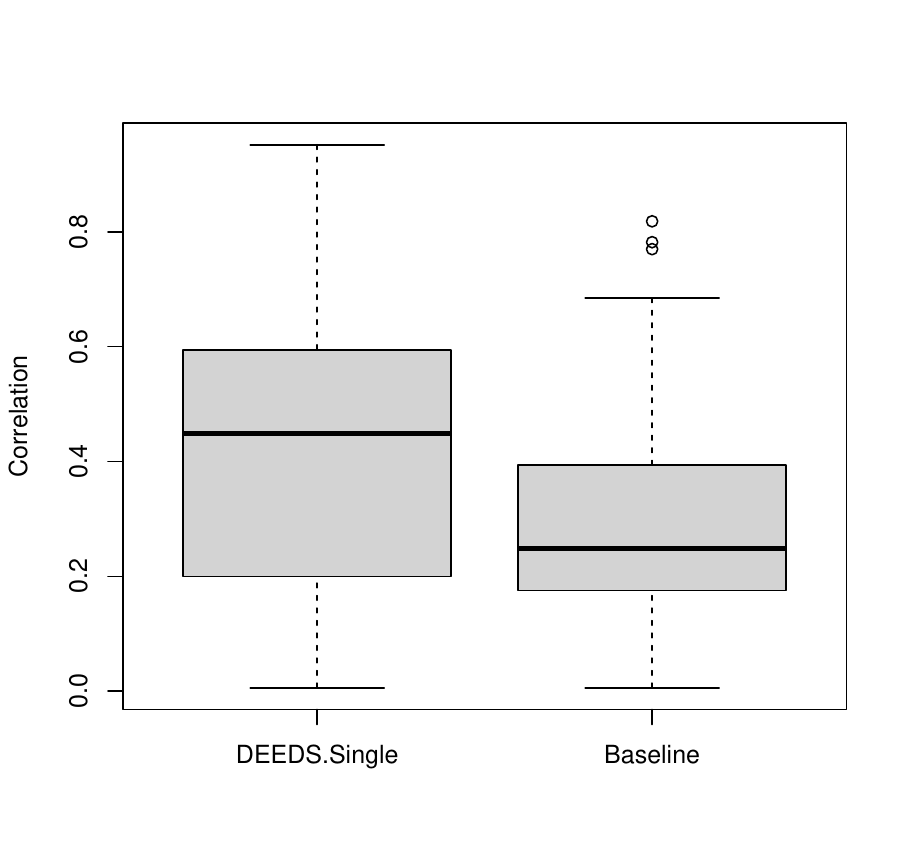}
\caption{Box plots of the correlation coefficients (in absolute terms) of the estimated rank orders of sets of 10 documents and their true rank orders, replicated 100 times.  The first plot corresponds to the DEEDS-single corpus, and the second to the baseline (random).}
\label{boxplotSingle}
\end{center}
\end{figure}

Regarding the DEEDS-single corpus, the TempSeq did not perform as well as for the temporally DEEDS-conflated collection. Although statistically significantly better than the baseline correlations (for tests based on Wilcox rank sum and the t-test, $\mbox{p-value}< 5.5\times 10^{-7}$), the median correlation coefficient between the estimated and the true rank orders was only 0.45 (see figure \ref{boxplotSingle}). This raises the question of why TempSeq under-performed on the DEEDS-single collection as compared to the DEEDS-conflated collection. These two collections differ in that the latter was created by merging all DEEDS documents for a given year, such that the DEEDS-conflated documents had a mean of 11,000 words as compared to only 175 words for the DEEDS-single collection. As such, the temporal sequencing for the DEEDS-single collection according to equation \ref{sigmahat} is based on a very small sample of words. Given the requirement that each word should occur at least once in two separate documents within a set, there are correspondingly fewer words informing the estimation of temporal order. 

As stated in section~\ref{intro}, the TempSeq approach for document sequencing allows the identification of words that are most informative for determining the correct temporal order. The process that we used to identify and analyze the informative words is as follows: As an example, we first considered a set of 10 documents from the DEEDS-conflated collection for which the TempSeq prediction of ordering, $\hat{\sigma}(l)$, resulted in a Spearman correlation coefficient of 0.92. Although the total number of unique words across these 10 documents was 15,757, when we considered the number of words that occurred at least once in two documents, the number of the relevant unique words declined to 5,988.  Of these 5,988 words, we selected those with frequency values in the top 50\textsuperscript{th} percentile. For each of these selected words, we computed the rule-of-thumb optimal bandwidth values under the predicted TempSeq order, $h_{amise,w,\hat{\sigma}(l)}$. On the basis of these optimal bandwidths, we extracted words that were in the top 88\textsuperscript{th} percentile for their maximum probability of occurrence score over the temporal rank domain in equation~\ref{prob}. The first condition ensured that the words under consideration were those that occurred with sufficient frequency (exceeding the median).  The second condition ensured that when a set of documents was listed in its correct temporal order, the informative words (i.e., the words that were most useful for attaining the correct temporal ordering using TempSeq method) were those whose occurrence clustered in near-by time periods. In turn, these clusterings resulted in higher word probability occurrence around those same time periods (refer to equation~\ref{prob}) as compared to the other time periods. Using the above word filtering procedure, we obtained a final list of around 360 relevant words. 

To interpret meaningfully the 360 relevant words in the list, we compared them with the representative topic terms from an LDA (Latent Dirichlet Allocation, ~\cite{blei2003latent}) topic modelling run on eight topics on the entire DEEDS-single documents~\citep{gervers2023topic}. In the pre-processing stage (prior to running the LDA algorithm), each document was split into tri-gram words (sequences of three consecutive words). The topic proportions for each of the documents were aggregated in accordance with their date of issue (see Figure 6.1 in ~\cite{gervers2023topic}). When we examined the dominant topics corresponding to the dates from the set of the 10 documents, we found that the vast majority of the informative words extracted using the procedure outlined above matched the words in the tri-gram topic terms. Some examples of the informative words and their contexts include \textit{finalis facta concordia} (indicating that the document is a 'final concordance made'),  \textit{anno regni regis} ('in the year of the king's reign'),  \textit{scripto sigillum meum} ('marking the document (with) my seal', indicating the sealing of a grant), \textit{pro omni seruitio} ('for all service', indicating that a transfer was not a simple donation), 
and \textit{perpetuam elemosinam} ('in perpetual alms', indicating that the transfer was a donation). 

\begin{figure}[!htb]
\begin{center}
\includegraphics[ width=12.0cm,height=9cm]{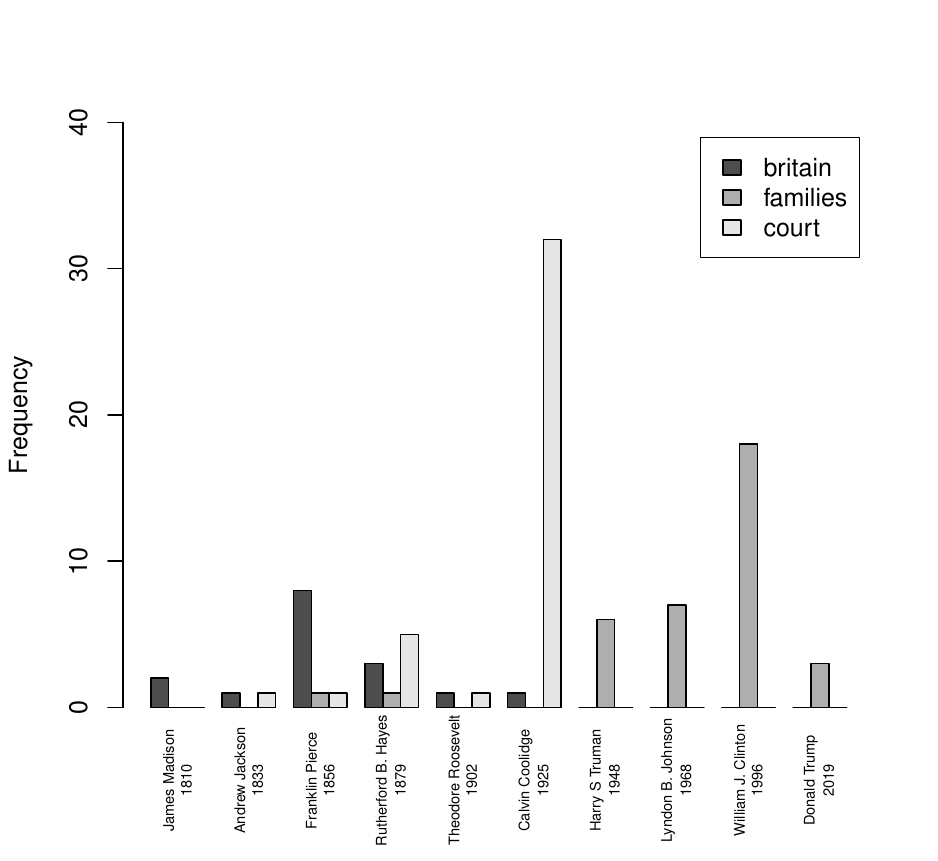}
\caption{A bar graph of the frequencies of the usage of the words \textit{Britain}, \textit{Families} and \textit{Court} in each presidential speech, indicated by year.}
\label{presidentwords}
\end{center}
\end{figure}

For the SOTU corpus, we similarly selected a set of 10 documents, each one separated by 23 years during the interval from 1810 to 2019. The Spearman correlation coefficient obtained between the TempSeq ordering method and the true orders for this set of documents was 0.95. The informative words were extracted and examined in the same way as described for the DEEDS-conflated corpus. Of a total of 29,537 words, there were 3,957 words that occurred at least once in at least two documents, of which the top 20\textsuperscript{th} percentile of the maximum probability of occurrence score over the temporal rank domain in equation~\ref{prob} yielded 435 words.  For the purposes of illustration, we present three words from the short list of relevant words: \textit{Britain}, \textit{Families}, and \textit{Court}. A bar graph of the frequencies of these words, counted from the selected set of 10 presidential speeches, is illustrated in figure~\ref{presidentwords} . 

We then ran an LDA topic modelling with five topics, where in the pre-processing stages, documents were split into bi-grams (sequences of two consecutive words), over the entire SOTU corpus.  
To enable an interpretation of the words that drove the temporal ordering, we examined the high ranking (top 20) bi-gram words of a topic associated with fiscal and commercial interests of the US. Within this topic, \textit{Great Britain} was one of the high ranking words. One of the uni-gram forms of the term, \textit{Britain}, turned up in the list of the top relevant words. 
From the 1810 SOTU address by James Madison, the term \textit{Britain} was invoked in the context of the naval blockade suffered by the US when the Napoleonic Wars had spilled into the Atlantic. Twenty-three years later, \textit{Britain } was discussed in the context of final settlement on the US North-East boundary and navigational safety concerns; another twenty-three years later, in 1856, Franklin Pierce's address discussed Britain in the contexts of her desire to dominate the Panama routes (and US refusal thereof), rights to fisheries,  increasing trade between the US and British Provinces in North America, and maritime rights regarding immunity from seizure: `\textit{... the private property of subjects and citizens of a belligerent on the high seas ... by the public armed vessels of the other belligerent, except it be contraband.}' Britain was a subject in Ruthford Hayes' 1879 address regarding the settlement of a dispute over rights to fisheries in Canadian waters. In two of the subsequent presidential addresses from the 10 documents, Britain was mentioned once in each, and not mentioned thereafter.  Thus, the correct temporal order (and the TempSeq ordering method) of the 10 selected documents optimized a gradual pattern of change in the usage of the word \textit{Britain}. 

Another informative word to TempSeq analysis is \textit{Families}, although it was not ranked highly from LDA, either as a uni-gram or as a portion of a word in a bi-gram. In the set of 10 SOTU documents, the word \textit{Families} was barely mentioned prior to the address by Harry Truman in 1948. The frequent usage of that word appears in the later time periods. For example, in the 1948 Address, Harry Truman mentioned \textit{Families} in the contexts of a social safety net and policies aiming to raise the standard of living for ordinary Americans. For example, we note `\textit{public housing for low-income families}', the provision of price support for farm commodities to enable `\textit{farm families ... to catch up with the standards of living enjoyed in the cities}' and anti-inflation measures to fight the `\textit{undermining} [of] \textit{the living standards of millions of families}'. In Lyndon B. Johnson's 1968 address, the word \textit{Families} was invoked to boast about the increase in the wealth accumulation of `\textit{most American families}' as `\textit{more and more families own their own homes ... television sets}'.  He urged congress to authorize more money to allow `\textit{new housing units for low and middle-income families}' to be built in order for `\textit{... thousands of families to become homeowners, not rent-payers}'. In William Clinton's 1996 Address, the word \textit{Families} is invoked in the context of sheltering `\textit{working families}' from the effects of government cuts.  In spite of cuts and a shrinking government, Clinton states his belief in the possibilities of cultivating `\textit{stronger families}'. Further, he speaks of the challenge to `\textit{strengthen America's families}' and thanks his wife for having taught him `\textit{the importance of families and children}'. Clinton also challenges `\textit{America's families to work harder to stay together}' because `\textit{families who stay together not only do better economically, their children do better as well}'. Furthermore, the word \textit{Families} is invoked in the context of health insurance policies --`\textit{... over one million Americans in working families have lost their health insurance}'. The 2019 Address by Donald Trump invokes the word \textit{Families} in his statement `\textit{We passed a massive tax cut for working families}'. The word \textit{Families} is also mentioned in the context of victims of criminal violence whom Trump had met -- `\textit{I have gotten to know many wonderful angel moms, dads and families}'.  

Among the 10 selected presidential speeches, \textit{Court} was found to be an informative word, despite not being ranked highly by LDA. Examining the bar graph in figure~\ref{presidentwords}, there is an abundant usage of the word (34 times) in the 1925 address by Calvin Coolidge. Although occurring a few times in the other prior presidential speeches (except for that of James Madison), the word \textit{Court} did not occur after this date. Emerging as a great power after the end of World War I, the United States sought a foreign policy with global influence. Calvin Coolidge invoked the word \textit{Court} primarily in discussing his administration's support in joining the Permanent Court of International Justice, which had been set-up in 1922. In his speech, Coolidge encouraged the senate to support adherence to the Court by arguing that the United States' interests would not be negatively affected, for example in stating `\textit{... by supporting the court we do not assume any obligations under the league ...}';  `\textit{... the statute creating the court shall not be amended without consent ...}'; `\textit{No provision of the statute ...  give[s] [the] court any authority to be a political rather than a judicial court}', and `\textit{If we support the court, we can never be obliged to submit any case which involves our interests for its decision}'.

The word \textit{Court} was also invoked in the 1879 address by Rutherford B. Hayes, although to a lesser extent than compared to that of Calvin Coolidge. Here, the primary contexts of usage involved criminal offenses and court administration. In the context of criminal offences, there is the example of the practice of polygamy in Utah, which would no longer be defended under the constitutional guarantee of religious freedom, `\textit{The Supreme Court of the United States has decided the law to be within the legislative power of Congress}'. Another instance concerns the urgency to introduce a justice system to prosecute criminals in the newly acquired territory of Alaska: `\textit{bill authorizing ... detention of persons charged with criminal offenses, and providing for an appeal to United States courts ...}'. In the context of court administration, we find the phrases, `\textit{The business of the Supreme Court is at present largely in arrears}'; `\textit{... magistrates who compose the court can accomplish more than is now done}', and `\textit{... additional circuit judges and the creation of an intermediate court of errors and appeals, which shall relieve the Supreme Court of a part of its jurisdiction ... }'.

\clearpage

\section{Error Analysis}
\label{error}
\begin{figure}[!htb]
\begin{center}
\includegraphics[ width=10.0cm,height=7cm]{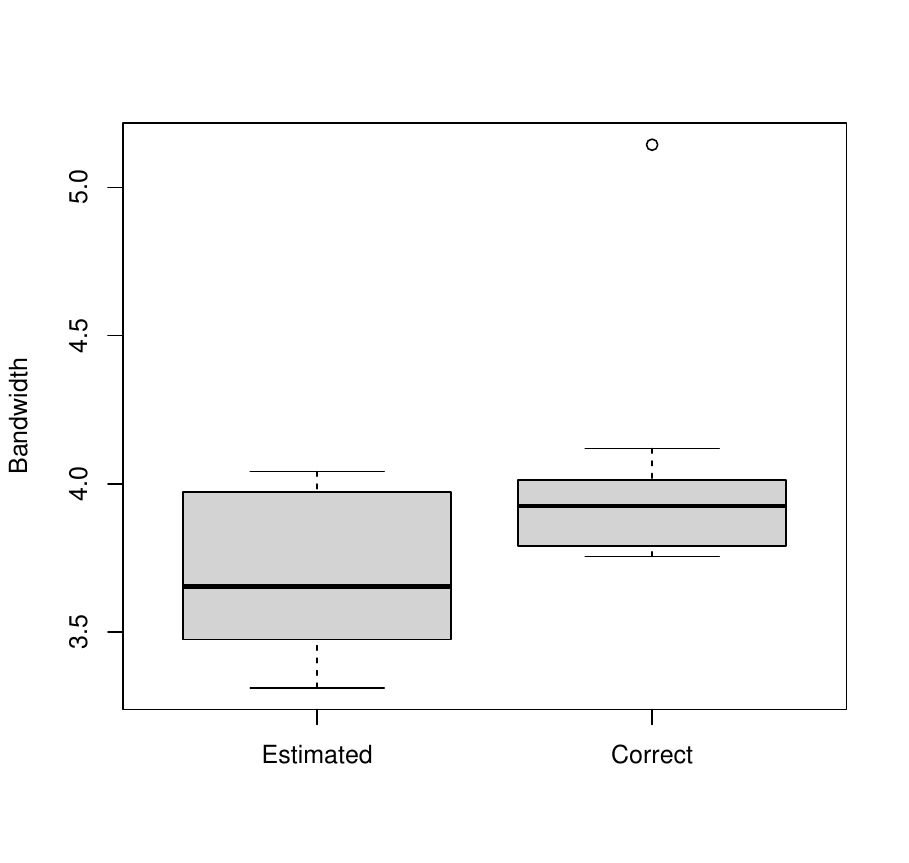}
\caption{On the left side is a box plot illustrating the bandwidth values obtained for the State of the Union (SOTU) documents with the lowest 10\ts{th} percentile correlations. On the right side is a corresponding box plot illustrating the bandwidth values obtained under the correct temporal ordering for the documents with the lowest 10\ts{th} percentile correlations.}
\label{boxplotSofUError}
\end{center}
\end{figure}

\begin{figure}[!htb]
\begin{center}
\includegraphics[ width=10.0cm,height=7cm]{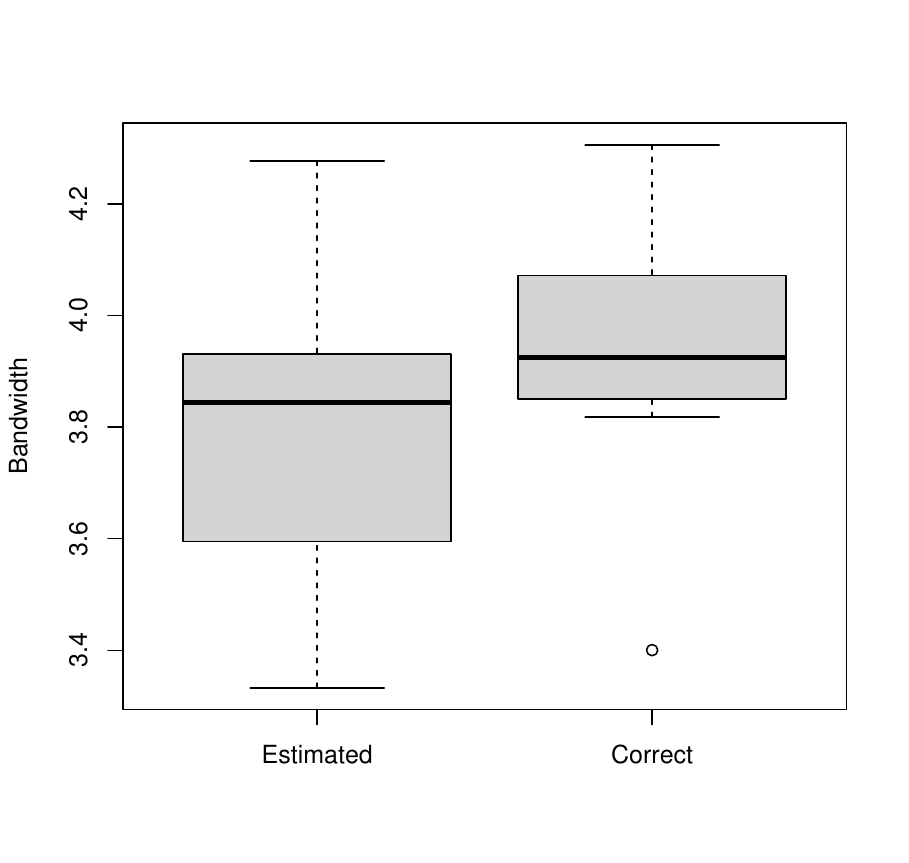}
\caption{On the left side is a box plot illustrating the bandwidth values obtained for the DEEDS-conflated documents with the lowest 10\ts{th} percentile correlations. On the right side is a box plot illustrating the bandwidth values obtained under the correct temporal ordering for the documents with the lowest 10\ts{th} percentile correlations.}
\label{boxplotDEEDSError}
\end{center}
\end{figure}

We conducted an error analysis on the subset of the 100 replication sets of 10 randomly selected documents for which TempSeq under-performed, with a cut-off of correlation coefficients falling below the 10\ts{th} percentile. For the SOTU corpus, these were the sets of documents for which correlation coefficient between estimated temporal ordering via the TempSeq method and their true temporal ordering was less than 0.27. For the richer DEEDS-conflated corpus, the corresponding threshold correlation was 0.62. When comparing the average bandwidth values, equation (\ref{sigmahat}), of the estimated temporal orderings for such sets of documents to that of the average bandwidth value under their correct temporal orderings, i.e., $H_{\sigma_{0}(l)}$, the values of $H_{\sigma_{0}(l)}$ were generally larger for the latter case, as shown in figures \ref{boxplotSofUError} and \ref{boxplotDEEDSError}. This reflects the lesser variability of word usage, and the gradual change in word frequency with time. The under-performance of TempSeq on the sets of documents under discussion is therefore explicable by the inadequate search runs of the Simulating Annealing algorithm that searches for the optimal temporal ordering in equation (\ref{sigmahat}).


\section{Conclusion}
\label{discussion}
A natural question that arises at this point is whether employing the models used in Large Language Models (LLM), which have shown unprecedented capabilities in understanding and generating natural language, can be used to solve the temporal sequential ordering problem described above. The fundamental problem with the possible application of LLM for the research question posed in this paper is that the amount of textual data required to train an LLM is typically massive (for example, the first BERT model was trained on over 3.3B words, \cite{devlin2018bert}). By comparison, the SOTU corpus has only a total of around 1.5M words, and the DEEDS corpus has only a total of around 2M words, with the dates of documents in each corpus spread across two and half centuries. Even to employ static word embeddings, which are ways to represent words as vectors in a multidimensional vector space, the smallest training sets required for reliable word representation using Word2Vec \citep{mikolov2013efficient} and GloVe \citep{pennington2014glove} were 24M words (a subset of Google News corpus) and 1B tokens (2010 Wikipedia), respectively. If pre-trained LLM models were to be leveraged for the task at hand, those models would need to have been trained on the right kinds of corpora, in the sense of topics, genre, time periods and language. Moreover, assuming we can obtain effective word representations using LLM, it is not clear how we could use them for the task of temporal ordering of a set of documents. More pertinently, the research question posed in this paper seeks to temporally order a set of documents (for example, sets of ten documents following the TempSeq experiments in sections~\ref{optimize} and \ref{results}) when a training corpus is not necessarily available, as is the case for many heritage texts. Far below the size of data required for LLM, a set of ten documents from SOTU has a total average count of 64K words (an average of 6400 words per Presidential Address), and a set of ten documents from DEEDS has a total average count of 1750 words (an average of 175 words per document).

Motivated by problems arising in the dating of historical and heritage texts, we set about in this paper to develop a method for assigning temporal rank orders to a sequence of documents when the date of issue is either missing or uncertain. In the historical context, the limited number and length of surviving manuscript texts presents a particular challenge in ordering. Our unsupervised method for document rank ordering relies on the principle that word usage changes gradually, on the scale of decades. Our method effectively captures changing word usage in the DEEDS and SOTU corpora by the bandwidth estimates. As shown (in section \ref{results}), the median of the correlation values for both corpora were significantly higher than the baseline from random ordering of sets of 10 documents. However, when the sizes of each of the documents are composed of fewer words, as in the case of the DEEDS-single, the TempSeq method doesn't perform as well due to the lack of adequate number of words on which to base the necessary estimates.

In practice, a reliable document rank ordering method should furnish the opportunity to identify particularly informative words for estimating the correct document orderings (for example, words with relatively high values of $h_{amise, w, \sigma(l)}$ for the optimal ordering). Equally, such a method provides us with the potential to identify anachronistic words, which may have been inserted into documents for nefarious reasons, such as the case of forged documents mentioned in section~\ref{intro}.

In our current experimental design, we selected orderings of documents that were approximately twenty years apart and extending over 200 years; our procedures for ordering performed significantly better than the randomization. In the future, we shall examine the performance of our method when selected documents are separated at variable time intervals.  We also plan to examine the temporal rank orderings of documents when the number of words in the documents is extremely limited, namely the particular case of Anglo-Saxon texts.

Acknowledgments: The authors gratefully acknowledge Prof. Paul Cumming of Bern University for his critical reading of this manuscript.

\bibliographystyle{plainnat}
\bibliography{main}

\appendix\footnotesize
\section{Annex 1}
\label{appendix}
\subsection{Non-parametric Kernel Regression for the Binomial Model}
\label{appendix1}
Suppose $(n_{w}(D_{i}), N(D_{i}), t_{D_{i}}), \; i=1, \ldots, n$ represents a sequence of data pairs, where $t_{D_{i}}$ represents the date of the ${D_{i}}^{{th}}$ document and $n$ denotes the size of the data; that is, the total number of documents in the collection.  Let {$n_{w}(D_{i})$} denote the number of occurrences of the word (or term) $w$ in document $D_{i}$. Let $N(D_{i})$ denote the total number of words (or terms) in document $D_{i}$. We are interested in modelling the probability of occurrence of the term {\em w} at time $t$. The Generalized Linear Models (GLM) for the binomial family is therefore a natural point of departure.

The GLM assume that the conditional likelihood of the response $Y$, given the explanatory variables $X$, has an exponential family form 
\begin{equation}
\label{expofam}
	\hspace{4cm} f(y\,|\,x)=\exp\left\{\frac{\theta(x)-b(\theta(x))}{a(\phi)} + c(y,\phi)\right\}
\end{equation} 
where $a(\cdot)$, $b(\cdot)$, and $c(\cdot, \cdot)$ are known functions, but the value of the {\it dispersion} parameter, $\phi$ is not necessarily known. The parameter $\theta(\cdot)$ is called the {\it canonical} parameter.  The conditional mean and variance of the above model can be shown to be 

		$$ m(x)=\textrm{E}(Y\,|\,X=x)=b^{\prime}(\theta(x)), $$
		and
		$$\textrm{Var}(Y\,|\,X=x)=a(\phi)b^{\prime\prime}(\theta(x))$$
		
In the parametric form of GLM, a function $g$ of the conditional expectation is regressed on the variable $X$  as 
    		$$g(m(x))= x^{t}\mbox{\boldmath{$\beta$}}$$
where {\boldmath{$\beta$}} is a vector of the regression coefficients. If $g=(b^{\prime})^{-1}$, then $g$ is designated as the {\it link} function because it links the conditional expectation to the canonical parameter $\theta$, such that we model $\theta(x)=x^{t}\beta$. Given independently observed data $\{(X_{i}, Y_{i}), i=1, \ldots, n \}$, the $\beta$ values are estimated by maximizing the joint conditional likelihood (in the form of equation (\ref{expofam}) or equivalently, by maximizing the joint conditional log-likelihood) over the $\beta$'s:
 \begin{eqnarray}
 \label{loglike}
 	\max_{\beta} L( \vec{y}\,|\, \vec{x}  )
	=\max_{\beta}\sum_{i=1}^{n}\log(f(y_{i}\,|\,x_{i}) )
 \end{eqnarray}	

In the binomial family setting, let $r$ be the number of trials and $Y$ the number of successes in the $r$ trials. Let $X$ be the predictor variable such that ${Y \sim \mbox{Bin}(r, \pi(X))}$, where $\pi(X)$ is the probability of success. Our interest is in estimating the mean of the sample proportion rather than the mean number of successes. Letting $Y^{*}=Y/r$, we wish to estimate ${ \pi(x)=\mbox{E}(Y^{*}\,|\,X=x,r)}$. Suppose {$\{(X_{i}, r_{i},Y_{i}), i=1, \ldots, n\}$} are samples drawn from { $(X,r,Y)$} where { $Y_{i}=r_{i}Y^{*}_{i} \sim\mbox{Bin}(r_{i}, \pi(X_{i}))$}. The form of the joint conditional log-likelihood can be written as 	

$$ L(Y^{*}_{1}, \ldots, Y^{*}_{n}\,|\,(X_{i}, r_{i}): i=1, \ldots, n) =\sum_{i=1}^{n} \left\{ \frac{  Y^{*}_{i}\theta(X_{i})-\log(1+\exp(\theta(X_{i})))  }{1/r_{i}} +\log \binom{{r_{i}}}{r_{i}Y^{*}_{i}}    \right\} $$
where {$\theta(X_{i})=\log\left( \frac{\pi(X_{i})}{(1-\pi(X_{i}))}\right)$, } {$b(\theta(X_{i}))=\log(1+\exp(\theta(X_{i})))$, $\phi_{i}=r_{i}$, $a(\phi_{i})=1/r_{i}$,} and {$c(Y^{*}_{i}, \phi_{i})=\log \binom{{r_{i}}}{r_{i}Y^{*}_{i}}$ }\citet{agresti2002categorical} . 

For the above binomial example, we model the canonical link function $\theta(x)=g(x, \mbox{\boldmath$\beta$})$ as a polynomial of degree at most $p$ ($p\ll n$) in the predictor variable, and {\boldmath$\beta$} is the vector of coefficients of this polynomial. In viewing the above as a reformulation of equation ($\ref{loglike}$), we wish to maximize (with respect to the {\boldmath$\beta$}'s) 
\begin{equation}
 \label{loglikeglm }
\hspace{5cm} \sum_{i=1}^{n}\log\{f(Y_{i}|X_{i},g(X_{i}; \mbox{\boldmath$\beta$}) )\}.
 \end{equation}	
One of the deficiencies from which this model suffers is its lack of flexibility, namely that the optimal values of the {\boldmath$\beta$}'s are {\it global} --- a set of parameter values over the entire domain. In the context of the problem addressed in this paper, we do not have a pre-defined idea as to the number of parameters that are necessary to model the probability of occurrence of tokens via the canonical link function (the {\it logit}, $\theta(x)$) as it varies over a given range of time. Our aim is to model the probability {\it locally} --- that is, to relax the global polynomial assumption and to allow the {\boldmath$\beta$}'s to adjust locally within a small neighbourhood of the domain space \citep{fan1995local}.
	
The local modelling approach thus leads to the following new {\it local log-likelihood} objective function:
\begin{equation*}
\label{loglikeker}
 L( \mbox{\boldmath$\beta$}(x))=\sum_{i=1}^{n}\log\{f(Y_{i}|X_{i},g(X_{i}; \mbox{\boldmath$\beta$}) )\} K_{h}(X_{i}-x)
\end{equation*}	
where 
$$\hat{\mbox{\boldmath{$\beta$}}}(x)=\arg\max_{\beta}L( \mbox{\boldmath{$\beta$}}(x)).$$
We define $K_{h}(u)\equiv h^{-1}K(u/h)$ where $K$ is a kernel function and the scaled factor $h$ is the associated bandwidth. The kernel function $K$ is typically a continuous, unimodal, symmetric, and non-negative. It satisfies the condition $\int_{-\infty}^{\infty}K(x)dx =1$ and decays fast enough to eliminate the contributions of remote data points. The kernel function could be a Gaussian distribution among many other possibilities, although in this paper, we used the t-distribution function with a low degree of freedom value (equal to 5), so as not excessively to discount distant data points. As a weight term, $K_{h}$ fits a polynomial regression curve around the data in the neighbourhood of $x$, where $h$ is the size of the local neighbourhood.

We build locally flexible {\boldmath$\beta$}'s at $u$ in the neighbourhood of the point $x$ for the canonical link function using the following expansion:
\begin{eqnarray}
\label{tayloexp}
\theta(u)=g(u, \mbox{\boldmath$\beta$}) &\approx& \beta_{0}(x)+\beta_{1}(x)(u-x)+\cdots +\beta_{p}(x)(u-x)^{p} 
\end{eqnarray}
where $\beta_{j}(x)=\frac{\theta^{(j)}(x)}{j!}$. Maximizing with respect to the {\boldmath$\beta$}'s, when the polynomial order is $p=0$, which is to say the {\it locally-constant} regression case, we obtain
\begin{eqnarray*}
	L( \mbox{\boldmath$\beta$}(x))& = &\sum_{i=1}^{n} \left[Y_{i}\beta_{0}(x)-r_{i}b(\beta_{0}(x)) \right. \\
	& & \;\;\;\; \left. +\log {\binom{r_{i}} {Y_{i}}}   \right] K_{h}(X_{i}-x).
\end{eqnarray*} 
Allowing $\hat{\beta}_{0}(x)=\hat{\theta}_{h}(x)$ to maximize the above expression, we find that
 \begin{eqnarray}
  \label{probsp0}
   \hat{\pi}(x) \equiv \hat{\pi}_{h}(x)&=& \exp(\hat{\beta}_{0}(x))/( 1+\exp(\hat{\beta}_{0}(x))) \nonumber \\
   &=& \sum_{i=1}^{n}\frac{Y_{i}K_{h}(X_{i}-x)}{r_{i}K_{h}(X_{i}-x)}
\end{eqnarray}
where $\hat{\pi}(x)$ is an estimate of the probability of success at $x$. When $p=1$, it follows
\begin{eqnarray*}
	L( \mbox{\boldmath$\beta$}(x)) &= &\sum_{i=1}^{n} \left[Y_{i}(\beta_{0}(x)+\beta_{1}(x)(X_{i}-x)) \right.\\
	& &\left. \;\;\;\; -r_{i}
	b(\beta_{0}(x)+\beta_{1}(x)(X_{i}-x)) \right. \\
	& &\;\;\;\; \left. +\log {\binom{r_{i}} {Y_{i}}}   \right] K_{h}(X_{i}-x)
\end{eqnarray*}
where $\hat{\beta}_{0}(x)$ and $\hat{\beta}_{1}(x)$ maximize the above equation. The maximizers can be found using numerical methods, such as that of Newton-Raphson, where the initial value of $\beta_{0}(x)$ is set to be the solution for the local polynomial estimator $p=0$ together with $\beta_{1}(x)=0$. The estimator $\hat{\pi}(x)$ (which doesn't have a closed form solution) is given by 
\begin{eqnarray}
    \label{probsp1}
    \hat{\pi}(x) \equiv \hat{\pi}_{h}(x)= \exp(\hat{\beta}_{0}(x)) /(1+\exp(\hat{\beta}_{0}(x)))
\end{eqnarray}

\section{Annex 2}
\label{appendix2}
\subsection{Bandwidth Estimation}
\label{appendix2a}
Regarding the notations used in this section, refer to Annex~\ref{appendix}.

\citet{fan1995local} derive a rule-of-thumb bandwidth parameter $h$ estimate for the curve $\hat{\pi}_{h}(x)$. Recalling that  ${\theta(X_{i})=\log\left( \frac{\pi(X_{i})}{(1-\pi(X_{i}))}\right) }$ (Annex~\ref{appendix}, section~\ref{appendix1}), the curve estimate $\hat{\pi}_{h}(x)$
was determined having first estimated the canonical {\em logit} function $\hat{\theta}_{h}(x)$ where the local polynomial fitting is for $p=1$, in equation \ref{tayloexp} of Annex~\ref{appendix}, section~\ref{appendix1}.  
The error incurred when estimating $\theta(x)$ with $\hat{\theta}_{h}(x)$ is measured using the asymptotic mean squared error (AMSE) criterion:
 
 \noindent When $h \rightarrow 0$ and $nh \rightarrow \infty$, as $n \rightarrow \infty$,
{ \begin{eqnarray}
\mbox{AMSE}\{\hat{\theta}_{h}(x)\}
&=&\left\{\int z^{2}K(z)dz \; \times\; \frac{\theta^{(2)}(x)}{2!}h^{2}\right\}^{2} \nonumber \\
& &+ \sigma^{2}(x; K) n^{-1}h^{-1} \nonumber
\end{eqnarray}}
where 
\begin{eqnarray}
 \label{var}
    \sigma^{2}(x; K)=\mbox{var}(Y|X=x)^{-1} f(x)^{-1} \times \int K^{2}(z)dz.
\end{eqnarray}
 
The expansion of $\mbox{AMSE}\{\hat{\theta}_{h}(x)\}$ above is split into the squared bias and variance terms of $\hat{\theta}_{h}(x)$, reflecting the bias and variance trade-offs. As the asymptotic expansion shows, low values of the bandwidth parameter $h$ decrease the bias at the cost of a high variance (insufficient smoothing). We also note that sparser regions of the design density $f(x)$ result in larger variance of the estimator. The unknown terms, such as $\theta^{(2)}(x)$ and $\sigma^{2}(x; K)$ would still need to be estimated. A convenient approach is rather to approximate the error for $\hat{\theta}_{h}(x) $ via the asymptotic mean integrated squared error (AMISE) defined to be 
 $$ \mbox{AMISE}\{\hat{\theta}_{h}(x)\}=\int \mbox{AMSE}\{\hat{\theta}_{h}(x)\}f(x)w(x)dx$$
where the design density $f$ and the weight function $w$ are included for stability reasons. According to this error criterion, the optimal bandwidth is given by:
 $$h_{amise}=(A/B)^{1/5}$$
 where $A= \int \sigma^{2}(x; K) f(x)w(x)dx$ and 
 { \small \begin{eqnarray}
 B= \left\{ \int z^{2} K(z)dz\right \}^{2} 
\times \left\{ \int \theta^{(2)}(x)^2 f(x) w(x) dx \right \} n  
 \end{eqnarray}}

\noindent The unknown quantities, $\sigma^{2}(x; K)$ and $\theta(x)$ can be estimated by fitting a $q$th-degree polynomial parametric fit where $q \geq p+1=2$. The estimated bandwidth $h_{amise}$ provides us with a rough and quick approach to calculating a bandwidth value to use in practice.

For theoretical results related to the estimator of $\hat{\pi}_{h}(x)$, such as the asymptotic distribution when the bandwidth $h \rightarrow 0$ and $nh \rightarrow \infty$ (thus allowing us to create a confidence band around the estimator), and for the form of the bias and variance of the estimator when $x$ is an interior and a boundary point, refer to \citet{fan1995local}.

\subsection{Bandwidth Computation}
\label{appendix2b}
All the computations were performed using the R language and environment for statistical computing,~\cite{R}. For codes, refer to \url{https://github.com/gitgelila/TempSeq}.

The bandwidth value $h_{amise,w,\sigma(l)}$ from section~\ref{optimize} is estimated following the above rule-of-thumb procedure to produce $h_{amise}$. To compute $h_{amise,w,\sigma(l)}$, we first need to estimate the unknown quantities $\sigma^{2}(x; K)$ and $\theta(x)$. For a particular permutation, $\sigma(l)$, of the true temporal sequence $l=(1,2, \ldots, 10)$ of a set of 10 documents, and on which the TempSeq method is to be run (see section~\ref{optimize}), the data has the form
$$\{(n_{w}(D_{\sigma(l)_{i}}), N(D_{\sigma(l)_{i}}), t(D_{\sigma(l)_{i}})), i=1, \ldots, 10\}.$$ The notation $\sigma(l)_{i}$ identifies the {\it i}\textsuperscript{th} document after permutation. $n_{w}(D_{\sigma(l)_{i}})$ counts the number of occurrences of the word $w$ in document $D_{\sigma(l)_{i}}$ which has a total of $N(D_{\sigma(l)_{i}})$ number of words. $t(D_{\sigma(l)_{i}})$ is the temporal rank of the date of issue of document $D_{\sigma(l)_{i}}$.  

Using the \textit{glm} (generalized linear model) function from the R statistical package, a parametric second degree polynomial logistic regression was fit to $\theta(x)$. This fit was used to estimate $\theta^{(2)}(x)$ and also $\sigma^{2}(x; K)$. The kernel function $K(x)$ is the Student's t-density function with degree of freedom equal to 5. The weight term $w(x)\geq 0$ was set to equal $1/10$ at each of the $10$ temporal positions of the independent variable, and zero elsewhere. The term $\int K^{2}(z)dz$ (in equation~\ref{var}) was numerically computed by randomly drawing samples from the Student's t-distribution with 5 degrees of freedom. The second moment of the Student's t-distribution with 5 degrees of freedom, $\int z^{2} K(z)dz = 5/3$.

\end{document}